\documentclass[conference]{IEEEtran}

\usepackage{cite}
\usepackage[numbers,sort&compress]{natbib}

\usepackage[utf8]{inputenc}
\usepackage{amsmath,amssymb}
\usepackage{graphicx}
\usepackage{bm}
\usepackage{enumerate}
\usepackage{comment}
\usepackage{url}
\usepackage{color,soul}
\usepackage{subcaption}
\usepackage{float}
\usepackage{booktabs}
\usepackage{multirow}

\definecolor{light-gray}{gray}{0.8}

\usepackage{amsmath,amssymb,amsfonts}
\usepackage{algorithmic}
\usepackage{graphicx}
\usepackage{textcomp}
\usepackage{subcaption}

\def\BibTeX{{\rm B\kern-.05em{\sc i\kern-.025em b}\kern-.08em
    T\kern-.1667em\lower.7ex\hbox{E}\kern-.125emX}}

\makeatletter
\newcommand{\linebreakand}{%
  \end{@IEEEauthorhalign}
  \hfill\mbox{}\par
  \mbox{}\hfill\begin{@IEEEauthorhalign}
}
\usepackage[table,xcdraw]{xcolor}

\begin{document}

\title{Time Series Modeling for Heart Rate Prediction: From ARIMA to Transformers \\}

\author{
\small 

\begin{tabular}[t]{c@{\extracolsep{8em}}c} 

1\textsuperscript{st} Haowei Ni\textsuperscript{*} & 2\textsuperscript{nd} Shuchen Meng \\
\textit{Columbia University} & \textit{Central University of Finance and Economics} \\
New York, USA & Beijing, China \\
Corresponding author: hn2339@caa.columbia.edu & scmeng19@163.com \\

\\

3\textsuperscript{rd} Xieming Geng & 4\textsuperscript{th} Panfeng Li \\
\textit{Chongqing University} & \textit{University of Michigan} \\
Chongqing, China &  Ann Arbor, USA\\
xieminggeng@foxmail.com & pfli@umich.edu \\

\\

5\textsuperscript{th} Zhuoying Li & 6\textsuperscript{th} Xupeng Chen \\
\textit{Johns Hopkins University} & \textit{New York University} \\
Baltimore, USA & Brooklyn, USA \\
zli181@jhu.edu & xc1490@nyu.edu \\

\\

7\textsuperscript{th} Xiaotong Wang & 8\textsuperscript{th} Shiyao Zhang  \\
\textit{Indepent Researcher} & \textit{Cornell University} \\
Beijing, China & Ithaca, USA \\
sophiaxiaotongwang@gmail.com & sz566@cornell.edu \\

\end{tabular}
}

\maketitle

\begin{abstract}
Cardiovascular disease (CVD) is a leading cause of death globally, necessitating precise forecasting models for monitoring vital signs like heart rate, blood pressure, and ECG. Traditional models, such as ARIMA and Prophet, are limited by their need for manual parameter tuning and challenges in handling noisy, sparse, and highly variable medical data. This study investigates advanced deep learning models, including LSTM, and transformer-based architectures, for predicting heart rate time series from the MIT-BIH Database. Results demonstrate that deep learning models, particularly PatchTST, significantly outperform traditional models across multiple metrics, capturing complex patterns and dependencies more effectively. This research underscores the potential of deep learning to enhance patient monitoring and CVD management, suggesting substantial clinical benefits. Future work should extend these findings to larger, more diverse datasets and real-world clinical applications to further validate and optimize model performance.
\end{abstract}

\section{Introduction}

Cardiovascular disease is one of the top causes of death globally, claiming millions of lives annually. To prevent any unforeseen events during surgery or the post-operative recovery, it is imperative to create forecasting models that allow healthcare facilities to monitor patients' vital bio-metric signs, such as heart rate, blood pressure, and electrocardiogram (ECG). Various statistical and machine learning models, including ARIMA and Prophet, are frequently employed in these contexts due to their proficiency in handling time series data. However, these traditional models exhibit significant limitations that undermine their reliability and performance, particularly in managing the complexity and non-linearity inherent in cardiovascular conditions. Consequently, given the critical need for effective forecasting models in cardiovascular disease management and the inadequacies of traditional models, there is an urgent necessity for more advanced approaches. 

The traditional ARIMA model, derived from the Box-Jenkins methodology~\cite{box1976time}, is well-regarded for its ability to manage seasonal variations and non-stationary trends in time series forecasting. It encounters challenges with noisy and sparse datasets commonly found in disease progression studies. Its reliance on manual parameter adjustment can hinder performance, particularly in rapidly evolving scenarios such as cardiovascular interventions, where swift and precise adjustments are critical. On the other hand, Prophet offers greater flexibility and automation, accommodating seasonality and trend changes with minimal manual intervention. However, its performance is highly dependent on accurate hyperparameter tuning and is less suited for high-frequency data, such as patient vital signs recorded in seconds. Prophet also risks overfitting with small datasets and can be computationally intensive with large datasets or numerous changepoints, resulting in prolonged training times and increased resource consumption.

Recognizing the limitations of traditional models in handling large datasets with extended time windows, deep learning-based methods have gained popularity in forecasting. AI is being leveraged in numerous other fields, including image processing for better resolution, RNN from adder's perspective, algorithm for portfolio optimization, showcasing its versatility and effectiveness in diverse applications. However, these methods have not been applied to cardiovascular diseases, which present unique complexities. To address these challenges, more advanced models capable of handling the complexity and non-linearity of cardiovascular disease data are needed. Deep learning models, such as neural networks~\cite{xing2024predicting,yan2024survival,zhou2024predict}, convolutional neural networks (CNNs), recurrent neural networks (RNNs), and transformers~\cite{jin2024time,su2024large,xiao2024xtsformer,shi2024scaling}, are well-suited for processing large, complex datasets and offer a promising solution for improving cardiovascular disease forecasting.

Despite the limitations of traditional models, the academic literature has notably lacked comparative studies evaluating the performance of traditional models against deep learning counterparts in the domain of cardiovascular disease (CVD) forecasting. Addressing this gap is essential for the distinct strengths and weaknesses of these methodologies and underscoring the potential benefits of advanced models. This study systematically compares ARIMA and Prophet with a suite of deep learning models, aiming to identify more resilient forecasting approaches capable of handling the intricate dynamics of patient biometric and vital sign data. Moreover, this research delves into the broader implications of model performance, emphasizing the critical necessity of accurate predictions in optimizing patient health management and clinical interventions. This study's findings underscore the significant potential of transformer-based models in biomedical time series forecasting. Transformer-based models, such as PatchTST and iTransformer, showed exceptional accuracy in predicting heart rate dynamics, highlighting their ability to handle complex temporal dependencies and non-linear relationships more effectively than traditional models. Key contributions of this paper include bridging the in comparative, offering insights into advancements in biomedical forecasting models, and enriching discussions on tailored healthcare strategies.

\section{Methods}
This section states the data collection preparation stage and complete procedures for model building and evaluation. 

\subsection{Data Description and Preparation} 
The heart rate time series dataset used in this study was obtained from the MIT-BIH Database. This dataset contained heart rate measurements for four different individuals. For T1 and T2, each series included 1800 equally spaced instant heart rate readings (in beats per minute) obtained at intervals of 0.5 seconds. Fig. 1 presents the time series plot for T1, showing the heart rate measurements over the studied interval. T3 and T4 were obtained in the same way with 950 and 851 measurements each. Z-score analysis was utilized to identify potential outliers in the dataset. After a thorough investigation, it was determined that these outliers were not due to input errors; therefore, they were retained for further analysis. Min-max normalization was applied to all models to ensure that all values fall within the range of 0 to 1 and to standardize the input data scale across each model. This step is crucial when working with deep learning models because it facilitates faster convergence during training and enhances overall model performance. Normalizing the data helps mitigate the effects of varying feature scales, leading to improved numerical stability and more reliable results. 

\begin{figure}[htbp]
    \centering
    \includegraphics[width=\linewidth]{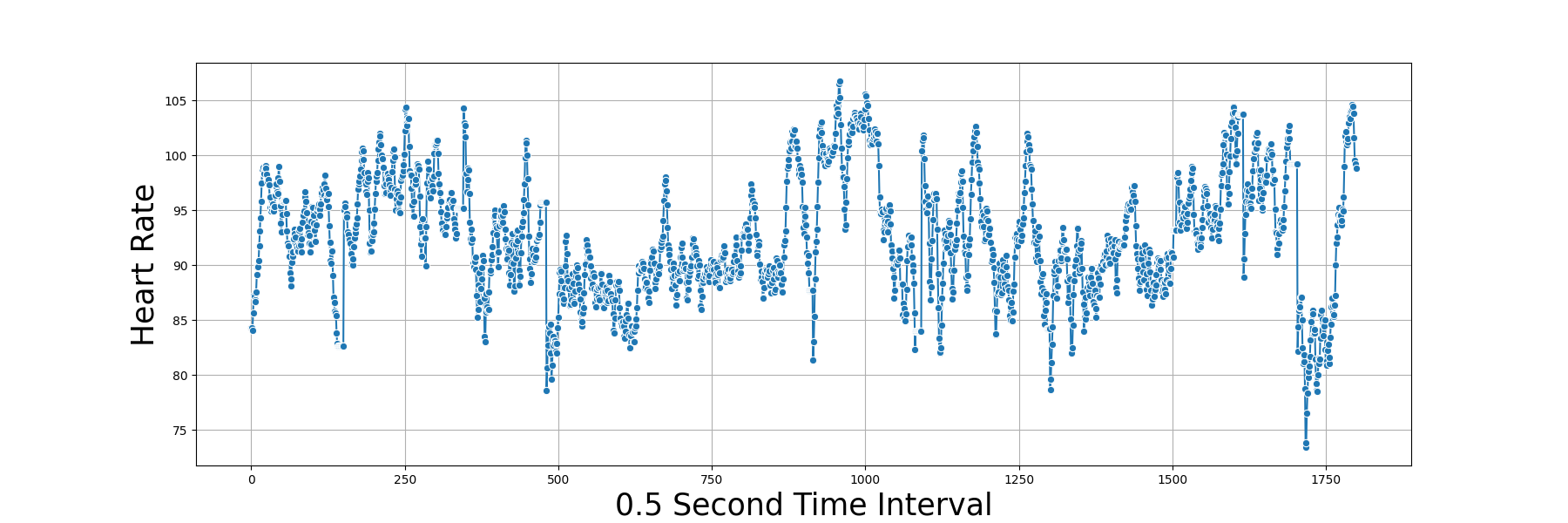}
    \caption{T1 Heart Rate time series.}
    \label{fig:time_series_heart_rate}
\end{figure}

\subsection{Framework} 
For each of the four-time series datasets (T1, T2, T3, T4), individual models were constructed using Seasonal-ARIMA, Prophet, LSTM, TSMixerx, TimesNet, TCN, PatchTST, and iTransformer. 

\subsection{Time Series Models Training} 
Seasonal ARIMA: This model combines both non-seasonal and seasonal elements, tailored specifically to capture the complex seasonal patterns in the heart rate datasets. It was used as a benchmark for evaluating the performance of other models.
Parameters such as \( p \) (non-seasonal autoregression order), \( d \) (non-seasonal differencing order), \( q \) (non-seasonal moving average order), \( P \) (seasonal autoregression order), \( D \) (seasonal differencing order), and \( Q \) (seasonal moving average order) were meticulously tuned. Additionally, \( S \) (seasonal period) was set according to the periodicity of the data. Model parameters were optimized using the AIC to ensure a balance between model simplicity and fit.

Prophet: We utilized the Facebook Prophet model as another benchmark. This model was employed with parameters such as \textit{changepoint\_prior\_scale}, \textit{seasonality\_prior\_scale}, and \textit{n\_changepoints} tuned through cross-validation. We also combined with adjustments aimed to optimize forecasting accuracy by controlling the model’s sensitivity to trend changes and seasonal fluctuations. Fig. 2 compares observed data points with the Prophet forecast, highlighting trend adjustments detection.

\begin{figure}[htbp]
    \centering
    \includegraphics[width=\linewidth]{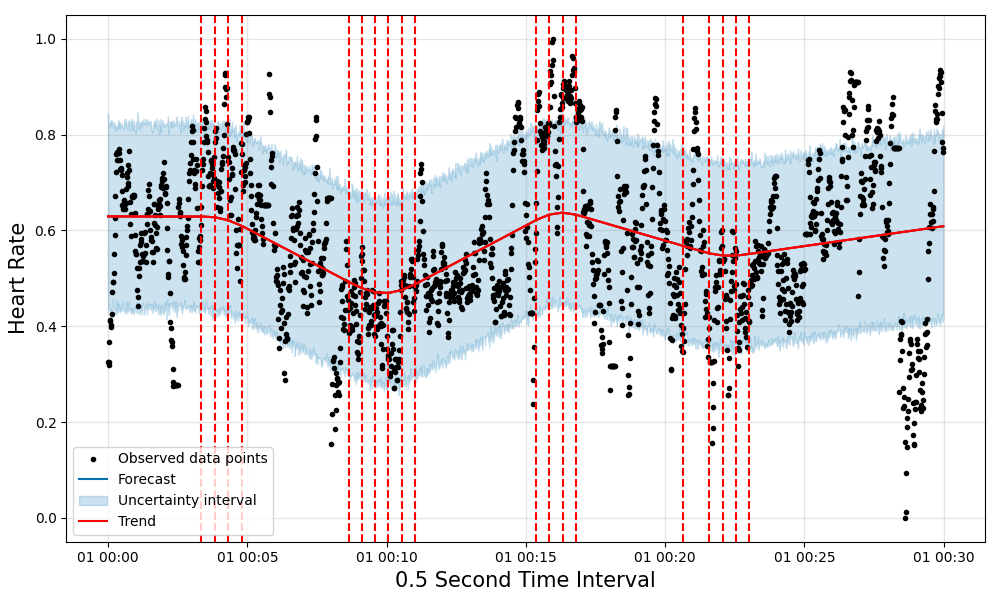}
    \caption{Prophet fit over the entire T1 dataset. The actual fit is depicted in light blue, the trend function is represented by a red line, and the trend changepoints are marked by red dashed vertical lines.}
    \label{fig:acf_pacf}
\end{figure}

\begin{table*}[htbp]
\centering
\caption{Performance Metrics of Averaged Time Series Models for Heart Rate Measurement Across Four Subjects}
\resizebox{\textwidth}{!}{%
\begin{tabular}{cc|cccccccc}
\toprule
\multicolumn{2}{c|}{\textbf{Metrics}} & \textbf{PatchTST} & \textbf{iTransformer} & \textbf{TCN} & \textbf{TSMixerx} & \textbf{LSTM} & \textbf{TimesNet} & \textbf{Prophet} & \textbf{SARIMA} \\
\midrule
\multirow{3}{*}{} & Avg MAE &  \textbf{1.993} & 2.746 & 3.476 & 3.493 & 3.693 & 3.657 & 4.104 & 5.591 \\
 & Avg MAPE &  \textbf{0.027} & 0.037 & 0.045 & 0.047 & 0.048 & 0.049 & 0.053 & 0.071 \\
 & Avg RMSE & \textbf{2.698} & 3.561 & 4.460 & 4.493 & 4.589 & 4.718 & 4.992 & 6.583 \\
\bottomrule
\end{tabular}%
}
\label{table:metrics}
\end{table*}


Long Short-Term Memory (LSTM) Network~\cite{li-stock-23,yan_21}: The LSTM model included two LSTM layers followed by a dense layer. Heart rate data was normalized using MinMaxScaler. We use three layers of LSTM with hidden dimension of 32. The model was trained using the MSE loss function to enhance performance.

Temporal Convolutional Network (TCN) has been used in time series prediction with huge success. We trained the model using dilated temporal convolutions to capture long-range dependencies in the heart rate data. Five layers of dilated convolutions with residual connections were employed to enhance training efficiency. The heart rate data was normalized before training. We set filter size to be three and dilation factor to be two. The MSE loss function used to guide the training process, ensuring effective learning of temporal patterns.

TSMixerx~\cite{ekambaram2023tsmixer}: For TSMixerx, we utilized a MLP block-based neural network. The model was trained on normalized heart rate data, with 5 layers of MLP layer and maximum feature dimension 16. The MSE loss function was used to optimize the model during training.

TimesNet~\cite{wu2022timesnet}: TimesNet was trained using a combination of convolutional layers to model both local and global temporal patterns by transforming the 1D time series into a set of 2D tensors. The model's architecture included 1 FFT block and 4 convolutional blocks. Normalized heart rate data was used, and the MSE loss function guided the training process to ensure accurate predictions.

PatchTST~\cite{nie2022time}: PatchTST involves dividing the input sequence into smaller patches and process each patch independently using transformer blocks. We set the patch size to be 12, with 6 transformer layers, and each layer with 8 heads. The heart rate data was normalized, and the model was trained using the MSE loss function. The self-attention mechanism within the transformer blocks allowed the model to capture intricate temporal dependencies.

iTransformer~\cite{chang-trans-24}: iTransformer was trained with improved positional encoding and attention mechanisms to better capture temporal dynamics. Instead of temporal attention, the model employs attention within the feature dimension. The model consisted of 4 transformer blocks, each has two transformer layers with 8 attention heads. The input heart rate data was normalized, and training was conducted using the MSE loss function to enhance predictive accuracy.

For deep learning models (LSTM, TCN, TSMixerx, TimesNet, PatchTST, and iTransformer), we used the Adam optimizer \cite{adam} with the following hyperparameters: learning rate (\( \text{lr} \)) = \( 10^{-3} \), \( \beta_1 \) = 0.9, \( \beta_2 \) = 0.999. The models were trained for 300 epochs with five-fold cross-validation. The learning rate was chosen based on preliminary experiments to ensure stable convergence and efficient training, while the Adam optimizer was selected for its ability to adapt the learning rate during training, which is beneficial for training deep neural networks.

\subsection{Performance Evaluation} 
In this study, three evaluation metrics were used to assess the models' performances, including Root Mean Square Error (RMSE), Mean Absolute Error (MAE), and Mean Absolute Percentage Error (MAPE). These metrics provided a comprehensive assessment of the model's predictive capability by quantifying the deviation between the forecasted and actual values. RMSE and MAE measure the absolute magnitude of the forecast errors, with RMSE giving higher weight to larger errors. MAPE calculates the average percentage difference between the forecasted and actual values, offering an intuitive measure of relative error. The evaluation metrics for each set of models were averaged across the four time series data to obtain a robust performance assessment.

\section{Experiments}
Table I showcases the obtained results after applying SARIMA and Prophet models as benchmark models alongside other deep learning models on the heart rate data. The evaluation results provide a complete insight into the performance of each model on forecasting the heart rate data by the second. The primary motivation behind this analysis is in assessing how well traditional statistical approaches compared to the more advanced deep learning techniques when used to monitor and forecast the patients' vital sign data. 

\subsection{Benchmark Model Performance} 
In this study, SARIMA and Prophet models were employed as baseline benchmarks for the evaluation of heart rate measurement data. Prophet generally outperformed SARIMA, as shown by lower values across all three metrics. These results indicate that while both SARIMA and Prophet can model heart rate data, Prophet provides more accurate predictions overall. However, both models struggled with the high variability and seasonality present in the datasets, reflected in their higher MAE and RMSE values.

\subsection{Deep Learning Based Model Performance} 
Several deep learning models were evaluated against the benchmark models. TCN and LSTM models showed significant improvements over SARIMA and Prophet. All the transformer-based models exhibit superior performance across three evaluation metrics. PatchTST achieves the best results, indicating its efficacy in modeling the intricate dynamics of heart rate data. iTransformer also shows notable improvements, reflecting the capability of transformers to handle temporal dependencies effectively. TimesNet and TSMixer, while not surpassing the top-performing models, still demonstrate notable improvements over the baseline models. 

\subsection{Interpretation of Results} 
The results above state a clear distinction in performance between traditional statistical models and deep learning models when applied to heart rate data. Although Prophet model shows slightly better results than SARIMA, it exhibited notably higher error rates than deep learning models. This suggests that traditional methods may struggle to capture the complex patterns and variability inherent in heart rate time series data. TCN and LSTM performed much better than the benchmarks, underscoring the advantage of applying deep learning models in this context. Transformer-based models like PatchTST and iTransformer, demonstrate superior performance. These models effectively capture temporal dependencies and nonlinear relationships, resulting in significantly lower MAE, MAPE, and RMSE values. The self-attention mechanism in transformer architectures allows these models to weigh the importance of different time points, leading to more accurate and robust predictions.
\begin{figure}[htbp]
    \centering
    \includegraphics[width=\linewidth]{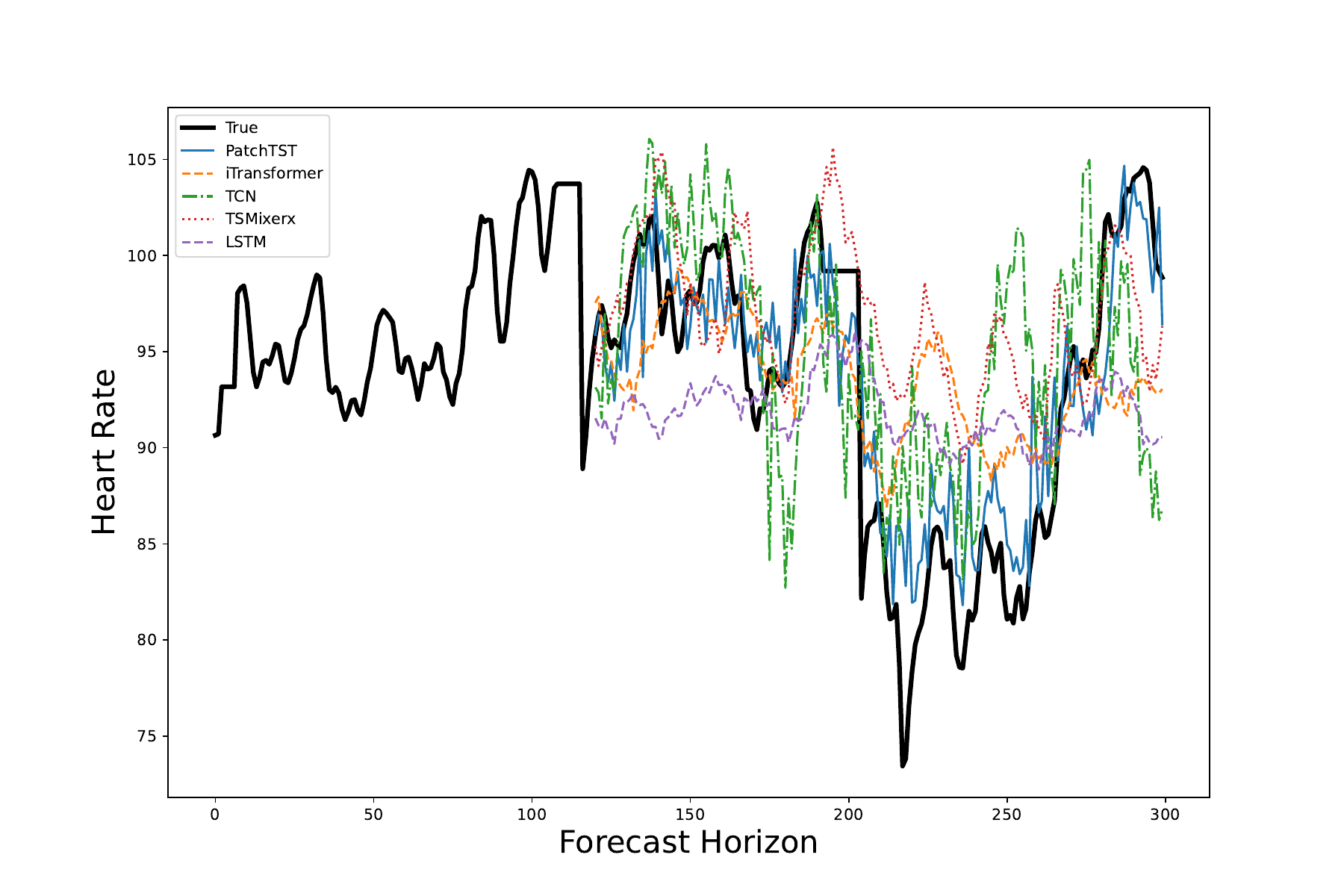}
    \caption{Comparison of True Values vs. Predicted Outputs from five best models}
    \label{fig:forecast_comparison}
\end{figure}
Fig. 3 comparing the true values against the predicted outputs from the five best models (PatchTST, iTransformer, TCN, TSMixerx, and LSTM) provides a visual representation of these results. It is evident from the graph that PatchTST and iTransformer closely follow the true values, indicating their high accuracy. TCN and TSMixerx also perform well, though with slightly more deviation from the true values. LSTM shows the most variability among the deep learning models, yet still provides a closer approximation to the true values than the traditional models. This visual comparison reinforces the quantitative findings, highlighting the effectiveness of transformer-based models and the overall superiority of deep learning approaches for heart rate time series forecasting.

\section{Discussion}

\subsection{Impact of Transformer-based models on Time Series Prediction Performance} 
Transformer-based models consistently outperformed traditional statistical models and other deep learning approaches. Among the transformer-based models, PatchTST demonstrated the best overall performance, effectively capturing the intricate dynamics of heart rate data. This superior performance can be attributed to the model's ability to handle complex temporal dependencies and non-linear relationships inherent in physiological data. The self-attention mechanism in transformer architectures allows these models to weigh the importance of different time points, leading to more accurate and robust predictions. This advancement sets a new standard for predictive performance in healthcare monitoring and other applications involving complex time series data.

\subsection{Theoretical and Practical Implications} 
The discoveries of this research carry significant ramifications for practical applications in clinical settings and the expansive scope of healthcare data analysis. This study has shown that advanced deep learning models, and in particular transformer-based architectures can achieve better levels of predictive accuracy than traditional time series statistical models. This study highlights the potential for these models to revolutionize prediction and monitoring in the era of continuous vital sign sensors. The implication is that healthcare facilities will be able to rely on better, timelier prediction of vital signs, which has great potential to transform clinical practice, making it easier to monitor patients, detect problems sooner, and intervene more quickly before adverse events occur.

\subsection{Study Limitations and Future Research Directions} 
It is worth noting the limitations of the presented study. Firstly, the dataset only consists of time series for four individuals with each time series varying between 7 and 15 minutes. As a consequence, the results of the study may not be generalizable to individuals with various syndromes or in different clinical settings. Secondly, the study only ran the models on the heart rate rather than other vital signs like blood pressure or levels of cholesterol. In terms of future studies, researchers can collect larger and more diverse datasets with other vital signs to enhance the model's generalizability and practical utility. Lastly, studying the long-term performance of these models and testing them in real-world clinical settings, for example, during surgery, post-surgery, or in an intensive care unit (ICU), will be needed to understand the utility and effectiveness of the study in monitoring patient health. 

\section{Conclusion}
This study successfully demonstrates the superiority of transformer-based models in forecasting cardiovascular health data by comparing them with traditional models like ARIMA and Prophet, as well as other deep learning approaches. Transformer-based models effectively capture the complex temporal dependencies and non-linear relationships inherent in heart rate data, resulting in substantially lower error metrics. The superior performance of these models demonstrates their potential to revolutionize predictive analytics in healthcare. 

The application of transformer-based models to heart rate data underscores their transformative potential in healthcare. These models' capacity to provide accurate predictions of vital signs enhances patient monitoring, enabling early detection of health issues and allowing for prompt medical interventions. This capability is crucial in clinical settings, where timely decisions can profoundly impact patient outcomes. Despite the limitations of this study, such as the small dataset and focus on heart rate alone, the findings highlight the potential of deep learning models in clinical settings.

In conclusion, this research underscores the significant improvements that transformer-based models bring to time series forecasting of heart rate data. The study sets a benchmark for future research and practical applications in health monitoring, advocating for the adoption of advanced deep learning techniques to achieve better predictive accuracy and pave the way for more sophisticated and effective healthcare solutions.

\renewcommand{\bibfont}{\footnotesize}

\footnotesize{
\bibliographystyle{IEEEtran}
\bibliography{main}

\begin{thebibliography}{10}
\providecommand{\url}[1]{#1}
\csname url@samestyle\endcsname
\providecommand{\newblock}{\relax}
\providecommand{\bibinfo}[2]{#2}
\providecommand{\BIBentrySTDinterwordspacing}{\spaceskip=0pt\relax}
\providecommand{\BIBentryALTinterwordstretchfactor}{4}
\providecommand{\BIBentryALTinterwordspacing}{\spaceskip=\fontdimen2\font plus
\BIBentryALTinterwordstretchfactor\fontdimen3\font minus \fontdimen4\font\relax}
\providecommand{\BIBforeignlanguage}[2]{{%
\expandafter\ifx\csname l@#1\endcsname\relax
\typeout{** WARNING: IEEEtran.bst: No hyphenation pattern has been}%
\typeout{** loaded for the language `#1'. Using the pattern for}%
\typeout{** the default language instead.}%
\else
\language=\csname l@#1\endcsname
\fi
#2}}
\providecommand{\BIBdecl}{\relax}
\BIBdecl

\bibitem{box1976time}
G.~E. Box, G.~M. Jenkins, G.~C. Reinsel, and G.~M. Ljung, \emph{Time series analysis: forecasting and control}.\hskip 1em plus 0.5em minus 0.4em\relax John Wiley \& Sons, 2015.

\bibitem{xing2024predicting}
Y.~Xing, C.~Yan, and C.~C. Xie, ``Predicting nvidia's next-day stock price: A comparative analysis of lstm, mlp, arima, and arima-garch models,'' \emph{arXiv preprint arXiv:2405.08284}, 2024.

\bibitem{yan2024survival}
X.~Yan, W.~Wang, M.~Xiao, Y.~Li, and M.~Gao, ``Survival prediction across diverse cancer types using neural networks,'' \emph{arXiv preprint arXiv:2404.08713}, 2024.

\bibitem{zhou2024predict}
C.~Zhou, Y.~Zhao, Y.~Zou, J.~Cao, W.~Fan, Y.~Zhao, and C.~Cheng, ``Predict click-through rates with deep interest network model in e-commerce advertising,'' \emph{arXiv preprint arXiv:2406.10239}, 2024.

\bibitem{jin2024time}
M.~Jin, H.~Tang, C.~Zhang, Q.~Yu, C.~Liu, S.~Zhu, Y.~Zhang, and M.~Du, ``Time series forecasting with llms: Understanding and enhancing model capabilities,'' \emph{arXiv preprint arXiv:2402.10835}, 2024.

\bibitem{su2024large}
J.~S. Su~et al., C.~Jiang, X.~Jin, Y.~Qiao, T.~Xiao, H.~Ma, R.~Wei, Z.~Jing, J.~Xu, and J.~Lin, ``Large language models for forecasting and anomaly detection: A systematic literature review,'' \emph{arXiv preprint arXiv:2402.10350}, 2024.

\bibitem{xiao2024xtsformer}
T.~Xiao~et al., ``Xtsformer: Cross-temporal-scale transformer for irregular time event prediction,'' \emph{arXiv preprint arXiv:2402.02258}, 2024.

\bibitem{shi2024scaling}
J.~Shi, Q.~Ma, H.~Ma, and L.~Li, ``Scaling law for time series forecasting,'' \emph{arXiv preprint arXiv:2405.15124}, 2024.

\bibitem{li-stock-23}
Z.~Li, H.~Yu, J.~Xu, J.~Liu, and Y.~Mo, ``Stock market analysis and prediction using lstm: A case study on technology stocks,'' \emph{Innovations in Applied Engineering and Technology}, pp. 1--6, 11 2023.

\bibitem{yan_21}
C.~Yan, Y.~Qiu, and Y.~Zhu, ``Predict oil production with lstm neural network,'' in \emph{Proceedings of the 9th International Conference on Computer Engineering and Networks}.\hskip 1em plus 0.5em minus 0.4em\relax Singapore: Springer Singapore, 2021, pp. 357--364.

\bibitem{ekambaram2023tsmixer}
V.~Ekambaram~et al., ``Tsmixer: Lightweight mlp-mixer model for multivariate time series forecasting,'' in \emph{ACM SIGKDD}, 2023.

\bibitem{wu2022timesnet}
H.~Wu~et al., ``Timesnet: Temporal 2d-variation modeling for general time series analysis,'' in \emph{The eleventh international conference on learning representations}, 2022.

\bibitem{nie2022time}
Y.~Nie, N.~H. Nguyen, P.~Sinthong, and J.~Kalagnanam, ``A time series is worth 64 words: Long-term forecasting with transformers,'' \emph{arXiv preprint arXiv:2211.14730}, 2022.

\bibitem{chang-trans-24}
P.~Chang~et al., ``A transformer-based diffusion probabilistic model for heart rate and blood pressure forecasting in intensive care unit,'' \emph{Computer Methods and Programs in Biomedicine}, vol. 246, 2024.

\bibitem{adam}
D.~P. Kingma and J.~Ba, ``Adam: A method for stochastic optimization,'' in \emph{International Conference on Learning Representations}, 2015.

\end{thebibliography}
}

\end{document}